%% file: main.tex
\documentclass{article} % For LaTeX2e
\usepackage{iclr2026_conference,times}

\input{math_commands.tex}

\usepackage{hyperref}
\usepackage{url}
\usepackage{booktabs}
\usepackage{graphicx}
\usepackage{caption}
\usepackage{subcaption}
\usepackage{multirow}
\usepackage{wrapfig}

\title{Supervised Metric Regularization Through Alternating Optimization for Multi-Regime Physics-Informed Neural Networks}

\author{Enzo Nicol\'as Spotorno, Josafat Leal Filho, \& Ant\^onio Augusto Fr\"ohlich \\
Department of Informatics and Statistics \\
Federal University of Santa Catarina \\
\texttt{\{enzoniko, josafat, guto\}@lisha.ufsc.br}
}

\iclrfinalcopy % Uncomment for camera-ready version

\begin{document}

\maketitle

\begin{abstract}
Standard Physics-Informed Neural Networks (PINNs) often face challenges when modeling parameterized dynamical systems with sharp regime transitions, such as bifurcations. In these scenarios, the continuous mapping from parameters to solutions can result in spectral bias or ``mode collapse'', where the network averages distinct physical behaviors. We propose a \textbf{Topology-Aware PINN (TAPINN)} that aims to mitigate this challenge by structuring the latent space via \textbf{Supervised Metric Regularization}. Unlike standard parametric PINNs that map physical parameters directly to solutions, our method conditions the solver on a latent state optimized to reflect the metric-based separation between regimes, showing $\approx 49\%$ lower physics residual (0.082 vs. 0.160). We train this architecture using a phase-based \textbf{Alternating Optimization (AO)} schedule to manage gradient conflicts between the metric and physics objectives. Preliminary experiments on the Duffing Oscillator demonstrate that while standard baselines suffer from spectral bias and high-capacity Hypernetworks overfit (memorizing data while violating physics), our approach achieves stable convergence with $\mathbf{2.18\times}$ lower gradient variance than a multi-output Sobolev Error baseline, and significantly fewer parameters than a hypernetwork-based alternative.

%COMMENT: are we sure they suffered from spectral bias?
\end{abstract}

\section{Introduction}

Physics-Informed Neural Networks (PINNs) have demonstrated significant potential for solving parameterized dynamical systems. However, modeling systems that exhibit qualitative behavioral changes, such as bifurcations from stability to chaos, remains a challenge. The literature suggests that standard Multi-Layer Perceptrons (MLPs) struggle to approximate the discontinuous or non-smooth dependence of the solution on the system parameters due to spectral bias \citep{rahaman2019spectral} and the singularity of the Jacobian at bifurcation points \citep{krishnapriyan2021characterizing}. While spectral bias reflects a representational limitation \citep{rahaman2019spectral}, ill-conditioned Jacobians near bifurcations primarily induce optimization pathologies \citep{krishnapriyan2021characterizing}. Recent works have also explored PINNs for bifurcation detection in nonlinear lattice systems \citep{shahab2025physics}. Nevertheless, existing solutions typically involve specialized architectures. \textbf{HyperPINNs} \citep{almeida2021hyperpinns} generate network weights conditioned on parameters to handle these transitions, while aiming to maintain parameter efficiency compared to training separate networks for each parameter. Other approaches, such as Mixture-of-Experts (MoE) \citep{bischof2022mixture}, explicitly route inputs to different sub-networks. These methods, however, introduce challenges such as routing instability (MoE) or indirect weight generation overhead (HyperPINNs).

In this work, we investigate a single-network architecture that incorporates \textbf{Supervised Metric Learning} (Triplet Loss) to regularize the latent space, and we name it \textbf{Topology-Aware PINN (TAPINN)}. We hypothesize that by explicitly enforcing a geometric structure in the latent space that mirrors the physical regimes, we can improve the trainability of the solver without the computational overhead of hypernetworks. To stabilize the training of these competing objectives (topology vs. physics), we employ an Alternating Optimization schedule.
We choose an LSTM-based encoder to capture temporal dependencies in short observation windows, which is crucial for distinguishing periodic from chaotic trajectories. Triplet loss is employed to cluster embeddings from trajectories in similar dynamical regimes while separating dissimilar ones, drawing from representation learning principles \citep{bengio2013representation} and manifold assumptions in high-dimensional data \citep{schroff2015facenet}. Triplets are formed in-batch using known forcing amplitudes $F_0$ as proxies for regime similarity (anchors/positives share the same $F_0$, negatives differ; no hard/semi-hard mining); we use Euclidean distance and margin = 0.2. This provides supervision without discrete regime labels. The alternating optimization schedule follows successful block-coordinate strategies in multi-objective PINNs \citep{wang2021understanding}, ensuring the latent manifold is stabilized before heavy physics enforcement and avoiding severe gradient conflicts. This phased approach is inspired by strategies for mitigating gradient pathologies in multi-objective PINNs \citep{wang2021understanding}.

%COMMENT: What does "without requiring discrete labels" mean?

\section{Methodology}
Below, we formalize the problem setting, describe the proposed encoder-generator architecture, and detail the loss functions and phased training procedure. We focus on a parameterized dynamical system governed by the residual $\mathcal{N}[\mathbf{x}(t); \lambda] = 0$. The objective is to reconstruct the full solution trajectory $\hat{\mathbf{x}}(t)$ given a short observation window $\mathbf{x}_{obs}$ associated with a parameter $\lambda$. Unlike standard Initial Value Problem (IVP) solvers that propagate forward from a single state $\mathbf{x}(0)$, our data-assimilation approach requires the encoder to process an observation window of the first 100 timesteps ($10\%$ of the trajectory) to accurately infer the underlying dynamical regime before reconstruction.
We specifically address the regime where $\nabla_\lambda \mathbf{x}$ may be ill-conditioned (e.g., near a bifurcation), making direct regression from $\lambda$ difficult for standard optimizers. Here, $\lambda$ is scalar (forcing amplitude $F_0$). We assume piecewise smoothness away from bifurcations. Unlike parametric baselines, $\lambda$ is used only indirectly via observed trajectories $\mathbf{x}_{obs}$.

TAPINN consists of an LSTM-based Encoder ($E$) and a PINN Generator ($G$). We use the term ``Topology-Aware'' to indicate that the method explicitly structures the metric geometry of the latent space to mirror separation between dynamical regimes (rather than computing strict topological invariants). The \textbf{Encoder:} $z = E(\mathbf{x}_{obs})$. Maps the input window to a latent vector $z$. The \textbf{Generator:} $\hat{\mathbf{x}}(t) = G(t, z)$, implemented as a 4-layer MLP with 32 hidden units and $\tanh$ activations. Unlike parametric baselines that require the known parameter $\lambda$, our method infers regime information solely from the short observation window $\mathbf{x}_{\rm obs}$ (first 100 timesteps), targeting realistic data-assimilation settings where $\lambda$ is unknown. LSTM is selected for its effectiveness in capturing long-term dependencies in dynamical trajectories; simpler alternatives (e.g., 1D CNNs) were explored but showed inferior regime discrimination in preliminary tests, therefore we left deeper experimentation for future work. We define a composite loss function $\mathcal{L}_{total} = \mathcal{L}_{data} + \alpha \mathcal{L}_{physics} + \beta \mathcal{L}_{metric}$. $\alpha$ and $\beta$ are tuned via grid search ($\alpha$=1.0, $\beta$=0.1 in our experiments); adaptive strategies remain future work \citep{wang2021understanding}.

To mitigate gradient conflicts observed when optimizing these terms simultaneously \citep{wang2021understanding}, we utilize a block-coordinate descent strategy. \textbf{Phase I (Metric Alignment)} optimizes the Encoder using a Triplet Loss $\mathcal{L}_{metric} = \max(0, d(z_a, z_p) - d(z_a, z_n) + m)$. This step organizes the latent space such that inputs from the same physical regime are clustered together. Intuitively, this geometric structuring forces the latent space to act as a linearized surrogate parameter manifold. By artificially grouping trajectories with similar forcing amplitudes and separating disparate ones, we reduce the effective condition number of the parameter-to-solution mapping, preventing the generator from collapsing into a spectral average of distinct periodic and chaotic regimes. This creates a more linearly separable representation, supporting manifold learning techniques like t-SNE for visualization \citep{maaten2008visualizing}. \textbf{Phase II (Physics Reconstruction)} optimizes the Generator on $\mathcal{L}_{physics}$ while freezing the Encoder, providing the solver with a stable conditioning variable $z$. 

\textbf{Interleaved Joint Tuning:} We train over 30 epochs as follows: Phase I (5 epochs, encoder-only on $\mathcal{L}_{metric}$), Phase II (20 epochs, generator-only on $\mathcal{L}_{physics} + \mathcal{L}_{data}$ with encoder frozen), then interleaved joint updates on $\mathcal{L}_{total}$ every $k=5$ batches ($\approx20\%$ of steps). This stabilizes the latent manifold early while keeping the encoder adaptable, preventing generator overfitting to a stale representation. By encouraging a latent space that linearizes the parameter-to-solution manifold, the approach may implicitly reduce the effective condition number of the Jacobian seen by the generator, potentially alleviating optimization difficulties near bifurcations identified in prior work \citep{krishnapriyan2021characterizing}.

\section{Experiments and Results}

We evaluate the method on the \textbf{Duffing Oscillator}, a canonical system exhibiting chaos. The system is governed by $\ddot{x} + \delta \dot{x} + \alpha x + \beta x^3 = F_0 \cos(\omega t),$ with standard parameters $\delta=0.3$, $\alpha=-1$, $\beta=1$, $\omega=1$. Varying the forcing amplitude $F_0 \in [0.3, 0.8]$ induces transitions from periodic to chaotic regimes. Such multi-scale dynamical systems are known to induce severe optimization pathologies in PINNs, as characterized by \citep{krishnapriyan2021characterizing}. While the Duffing system exhibits clear and relatively sharp regime transitions, subtler or higher-codimension bifurcations in systems such as the Lorenz attractor or reaction-diffusion equations (e.g., Allen-Cahn), broader validation is left for future work. Simulating regimes from Period-1 to Chaos by varying the forcing amplitude $F_0 \in [0.3, 0.8]$, we generated a dataset of 500 trajectories for each regime ($F_0 \in \{0.3, 0.5, 0.8\}$) using a 4th-order Runge-Kutta solver with $dt=0.01$. The observation window consists of the first 100 timesteps ($10\%$ of the trajectory). Models were trained for 30 epochs using the Adam optimizer with a learning rate of $10^{-3}$. 

We compare our approach against three baselines. \textbf{(I) Parametric Baseline:} A standard PINN, a 4-layer MLP with 64 hidden-units each, that receives the explicit parameter $\lambda$ as input ($Input: t, \lambda$). Unlike this and the HyperPINN baseline (both conditioned on known $\lambda$), TAPINN infers regime from partial observations $\mathbf{x}_{obs}$, targeting data-assimilation settings where $\lambda$ is unknown. \textbf{(II) HyperPINN Baseline:} A hypernetwork architecture where weights are predicted as functions of $\lambda$. This represents a high-capacity alternative for multi-regime modeling, designed for parameter efficiency  \citep{almeida2021hyperpinns}. \textbf{(III) Multi-Output Baseline:} Identical to our architecture (LSTM Encoder + Generator) but trained via \textbf{standard joint optimization} with a Sobolev ($H^1$) loss that enforces constraints on both $\mathbf{x}$ and $\dot{\mathbf{x}}$ \citep{czarnecki2017sobolev}. This isolates the combined effect of our Metric Regularization and Alternating Optimization against standard derivative-constrained training.

\begin{table}[t]
    \centering
    \caption{\textbf{Performance Comparison.} We report Physics Residual (mean squared ODE residual $\mathcal{L}_{physics} = \frac{1}{N_c}\sum \|\mathcal{N}[\hat{\mathbf{x}}(t_i); \lambda]\|^2_2$ over $N_c=10^4$ uniformly sampled collocation points $t_i \in [0,T]$; no additional weighting), Model Complexity (Exact Parameters), and Test MSE. Note that while HyperPINN achieves low Data MSE, its high Physics Residual indicates overfitting. Our method achieves the best physics compliance with $5\times$ fewer parameters than HyperPINN. All other models have a similar parameter count for fairness. Since the Parametric baseline doesn't use the encoder we matched its parameter count by increasing the MLP capacity.}
    \label{tab:results}
    \begin{tabular}{lrrr}
    \toprule
    \textbf{Method} & \textbf{Physics Res.} ($\downarrow$) & \textbf{Params} & \textbf{Data MSE} ($\downarrow$) \\
    \midrule
    Parametric Baseline & 0.160 & 8,577 & 0.392 \\
    Multi-Output & 0.192 & 8,069 & 0.426 \\
    HyperPINN & 0.158 & 39,169 & \textbf{0.281} \\
    \textbf{Ours (AO)} & \textbf{0.082} & 8,003 & 0.425 \\
    \bottomrule
    \end{tabular}
\end{table}

\begin{figure}[t]
    \centering
    \includegraphics[width=1.0\linewidth]{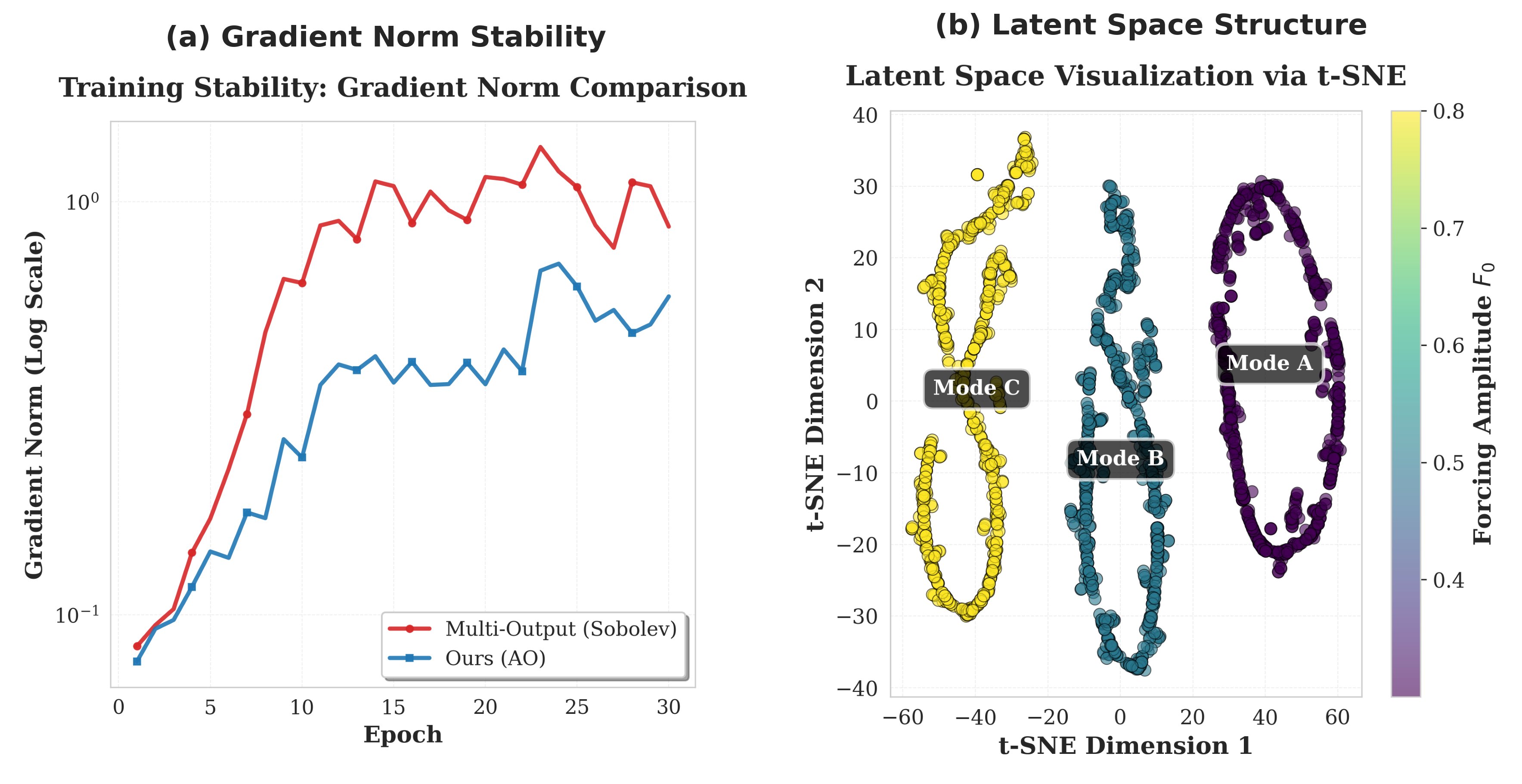}
    \caption{Qualitative analysis of the Topology-Aware training process: (a) Gradient norms during training. The Multi-Output baseline (orange) shows significant spikes near regime transitions compared to ours (blue). (b) t-SNE visualization of the learned embeddings $z$, colored by regime ($F_0$). Distinct clusters emerge without discrete supervision.}\vspace{-20pt}
    \label{fig:qualitative_analysis}
\end{figure}

In our experiments, the Multi-Output baseline exhibited severe numerical instability. While Sobolev training is effective for smooth functions, the inclusion of gradient constraints near the bifurcation point resulted in gradient norms $\mathbf{2.14\times}$ higher on average compared to our method, with a variance $\mathbf{2.18\times}$ larger (see Figure \ref{fig:qualitative_analysis}a). This suggests that enforcing high-order constraints on the encoder without explicit metric regularization leads to an unstructured latent space, exacerbating optimization pathologies near bifurcations. Moreover, we include HyperPINNs as a high-capacity benchmark. As shown in Table \ref{tab:results}, HyperPINN suffers from a ``memorization" pathology: it achieves the lowest Data MSE ($0.281$) but a high Physics Residual ($0.158$), indicating it overfits trajectory points without satisfying the governing ODE. One might question whether TAPINN's lower physics residual is merely a regularization artifact of its smaller capacity compared to HyperPINN (8,003 vs. 39,169 parameters). However, our ablation demonstrates otherwise. Both the Parametric and Multi-Output baselines operate at the same $\approx 8$k capacity but yield physics residuals approximately $2\times$ higher than our approach (0.160 and 0.192 vs. 0.082). This confirms that the topology-aware structuring, rather than merely restricted capacity, is the primary driver of the improved physical compliance.

In contrast, our Topology-Aware approach achieves the lowest physics error ($0.082$) while using substantially fewer parameters ($8,003$ vs $39,169$), demonstrating that structuring the latent space can be more parameter-efficient than brute-force weight generation. Beyond visual clustering, we quantify the structure of the learned latent space $z$ by training a linear probe to regress the physical parameter $F_0$ from $z$. Our method achieves a \textbf{Prognostics MSE} of $3.5 \times 10^{-4}$, confirming that the encoder learns a highly structured, linearizable representation of the chaotic regimes (see Figure \ref{fig:qualitative_analysis}b), which is not explicitly enforced by the physics loss alone. While this initial proof-of-concept focuses on characterizing the optimization stability and the resulting latent space structure, a rigorous comparison of the reconstructed trajectories against ground truth in physical coordinates remains a priority for extended validation. Finally, to isolate the contribution of our phased training strategy versus the metric loss itself, we compare our approach against a version trained with standard joint optimization (summing Metric, Physics, and Data losses). This ``Joint Training" configuration yielded a significantly higher physics residual ($\approx 0.158$), performing nearly identically to the standard baseline ($0.160$). This failure confirms that metric regularization alone is insufficient; the Alternating Optimization schedule is critical to resolve gradient conflicts, allowing the latent topology to stabilize before being subjected to the competing gradients of the physics constraints.

\section{Conclusion}
We presented an exploratory proof-of-concept Topology-Aware PINN framework that leverages Supervised Metric Regularization and Alternating Optimization to effectively address gradient conflicts and regime transitions in multi-regime dynamical systems. The encouraging preliminary results on the Duffing oscillator, demonstrating substantially improved physics compliance, training stability, and well-structured latent representations, highlight the promise of explicitly organizing the latent space. These findings naturally open several exciting avenues for future work, including theoretical analyses of Jacobian conditioning and spectral properties, rigorous statistical validation across random seeds, sensitivity studies on observation-window length, and broader empirical evaluation on PDE systems, continuous parameter regimes, and noisy data. In addition, comparisons with capacity-matched variants of high-parameter architectures such as HyperPINNs will provide deeper insight into the benefits of metric regularization. Lastly, future work will also focus on extensive benchmarking of TAPINN against domain decomposition methods \citep{jagtap2020extended} and operator learning frameworks \citep{li2020fourier}. We believe this lightweight approach offers a practical and efficient pathway toward more robust physics-informed modeling of complex dynamical systems and will be extended in subsequent research.

\subsubsection*{Acknowledgments}

This work was partially funded by Fundação de Apoio da UFMG (Fundep), through Linha VI – Conectividade 
Veicular, a prioritary program from Mover (Mobilidade Verde e Inovação), project AutoDL (29271.03.01/2023.04-00) and Auto5G (29271.02.01/2022.01-00).

\bibliography{references}
\bibliographystyle{iclr2026_conference}

\end{document}

%% file: math_commands.tex
\usepackage{amsmath,amsfonts,bm}

\def\eqref#1{equation~\ref{#1}}
\def\1{\bm{1}}

\DeclareMathAlphabet{\mathsfit}{\encodingdefault}{\sfdefault}{m}{sl}
\SetMathAlphabet{\mathsfit}{bold}{\encodingdefault}{\sfdefault}{bx}{n}

%% file: references.bib
@article{rahaman2019spectral,
  title={On the spectral bias of neural networks},
  author={Rahaman, Nasim and others},
  journal={International Conference on Machine Learning (ICML)},
  year={2019},
}

@article{krishnapriyan2021characterizing,
  title={Characterizing possible failure modes in physics-informed neural networks},
  author={Krishnapriyan, Aditi and others},
  journal={Advances in Neural Information Processing Systems (NeurIPS)},
  year={2021},
}

@article{almeida2021hyperpinns,
  title={{HyperPINNs}: Learning parameterized differential equations with physics-informed hypernetworks},
  author={Almeida, Diogo and others},
  journal={Advances in Neural Information Processing Systems (NeurIPS)},
  year={2021},
}

@article{czarnecki2017sobolev,
  title={{S}obolev training for neural networks},
  author={Czarnecki, Wojciech M and others},
  journal={Advances in Neural Information Processing Systems (NeurIPS)},
  year={2017},
}

@article{wang2021understanding,
  title={Understanding and mitigating gradient flow pathologies in physics-informed neural networks},
  author={Wang, Sifan and others},
  journal={SIAM Journal on Scientific Computing},
  year={2021},
}

@article{schroff2015facenet,
  title={{F}ace{N}et: A unified embedding for face recognition and clustering},
  author={Schroff, Florian and others},
  journal={Proceedings of the IEEE Conference on Computer Vision and Pattern Recognition (CVPR)},
  year={2015},
}

@article{jagtap2020extended,
  title={Extended physics-informed neural networks ({XPINNs})},
  author={Jagtap, Ameya D and Karniadakis, George Em},
  journal={Communications in Computational Physics},
  year={2020},
}

@article{li2020fourier,
  title={Fourier neural operator for parametric partial differential equations},
  author={Li, Zongyi and others},
  journal={International Conference on Learning Representations (ICLR)},
  year={2021},
}

@inproceedings{bischof2022mixture,
  title={Mixture-of-experts-ensemble meta-learning for physics-informed neural networks},
  author={Bischof, Rafael and Kraus, Michael A},
  booktitle={Proceedings of 33. forum bauinformatik},
  year={2022}
}

@article{bengio2013representation,
  title={Representation learning: A review and new perspectives},
  author={Bengio, Yoshua and Courville, Aaron and Vincent, Pascal},
  journal={IEEE Transactions on Pattern Analysis and Machine Intelligence},
  volume={35},
  number={8},
  pages={1798--1828},
  year={2013},
  publisher={IEEE}
}

@article{maaten2008visualizing,
  title={Visualizing data using t-{SNE}},
  author={Maaten, Laurens van der and Hinton, Geoffrey},
  journal={Journal of Machine Learning Research},
  volume={9},
  number={Nov},
  pages={2579--2605},
  year={2008}
}

@article{shahab2025physics,
  title={Physics-informed neural networks for high-dimensional solutions and snaking bifurcations in nonlinear lattices},
  author={Shahab, Muhammad Luthfi and Suheri, Fidya Almira and Kusdiantara, Rudy and Susanto, Hadi},
  journal={Physica D: Nonlinear Phenomena},
  pages={134836},
  year={2025},
  publisher={Elsevier}
}
